\documentclass[preprint,12pt]{elsarticle}

\usepackage{amssymb}

\usepackage{amsmath} 
\usepackage{amsfonts} 
\usepackage{multirow}
\usepackage{caption}
\usepackage{booktabs}
\usepackage{subcaption}

\journal{Neural Networks}

\begin{document}

\begin{frontmatter}

\title{Enhancing LLM Reasoning with Multi-Path Collaborative Reactive and Reflection agents}

\author[label1]{Chengbo He}
\author[label1]{Bochao Zou}
\author[label1]{Xin Li}
\author[label1]{Jiansheng Chen}
\author[label2]{Junliang Xing}
\author[label1]{Huimin Ma}

\begin{abstract}

Agents have demonstrated their potential in scientific reasoning tasks through large language models. However, they often face challenges such as insufficient accuracy and degeneration of thought when handling complex reasoning tasks, which impede their performance. To overcome these issues, we propose the Reactive and Reflection agents with Multi-Path Reasoning (RR-MP) Framework, aimed at enhancing the reasoning capabilities of LLMs. Our approach improves scientific reasoning accuracy by employing a multi-path reasoning mechanism where each path consists of a reactive agent and a reflection agent that collaborate to prevent degeneration of thought inherent in single-agent reliance. Additionally, the RR-MP framework does not require additional training; it utilizes multiple dialogue instances for each reasoning path and a separate summarizer to consolidate insights from all paths. This design integrates diverse perspectives and strengthens reasoning across each path. We conducted zero-shot and few-shot evaluations on tasks involving moral scenarios, college-level physics, and mathematics. Experimental results demonstrate that our method outperforms baseline approaches, highlighting the effectiveness and advantages of the RR-MP framework in managing complex scientific reasoning tasks.

\end{abstract}


\end{frontmatter}

\section{Introduction}

Large Language Models-based agents have demonstrated significant potential in scientific reasoning tasks. However, when faced with complex scientific reasoning challenges, these models often exhibit limited performance due to insufficient accuracy~\cite{feng2025retrieval,ZHANG2024107059,liang2023encouraging}. For instance, in tasks involving moral judgment or multi-level knowledge integration (such as university-level scientific problems), agents are capable of generating preliminary and comprehensible outputs but frequently struggle to provide comprehensive and accurate solutions~\cite{ma2023let, sel2024skin, madaan2024self}. Although step-by-step reasoning has somewhat enhanced the capabilities of agents~\cite{brown2020language}, fundamental issues such as hallucination persist when addressing these complex tasks, leading agents to generate content that appears reasonable but is inherently illogical~\cite{huang2023survey}.

Relevant studies have proposed solutions, among which self-correction, the simplest form of post-hoc adjustment, has garnered significant attention in recent years~\cite{pan2024automatically,lu2023self}. This approach leverages Large Language Models (LLMs) to generate feedback and optimize their own outputs, enabling LLMs to automatically rectify their generated content under zero-shot or few-shot prompts~\cite{jiang2024llms}. Although error detection is a prerequisite for self-correction, effectively implementing it remains a challenge. Previous research indicates that LLMs, similar to humans, do not always produce optimal outputs on the first attempt. Consequently, researchers have introduced the SELF-REFINE method, which assists agents in continuously improving their performance on specific tasks through iterative feedback and optimization~\cite{madaan2024self} .However, despite the potential of self-reflection to enhance answer quality, it relies on the LLM's self-assessment capabilities, which have yet to be fully validated~\cite{shinn2024reflexion}. Moreover, the reflection process of a single agent may lead to the Degeneration-of-Thought (DoT). Specifically, once a Large Language Model-based agent establishes confidence in its responses, it becomes incapable of generating novel insights through subsequent self-reflection, even if its initial stance is erroneous~\cite{liang2023encouraging}.

\begin{figure*}[t]
\centering
\includegraphics[width=\textwidth]{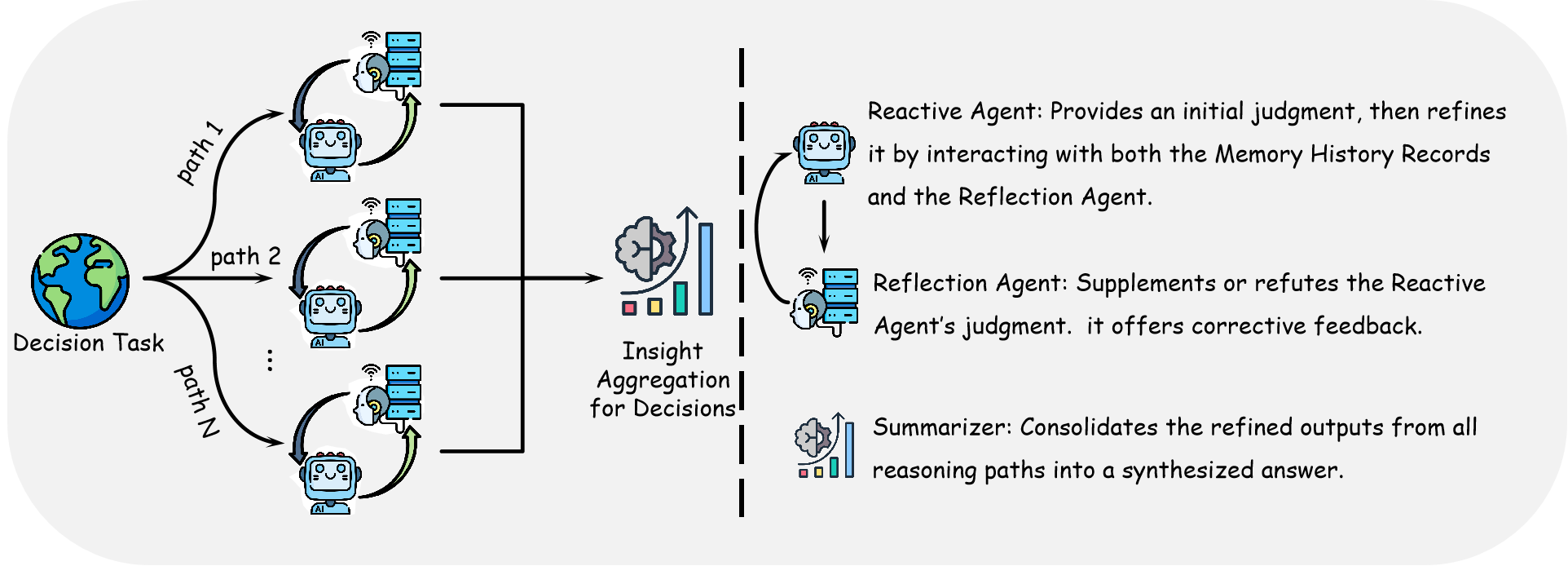} 
\caption{Reactive and Reflection agents with Multi-Path Reasoning }
\label{figure1} 
\end{figure*}

We propose the Reactive and Reflection agents with Multi-Path Reasoning (RR-MP) framework to address the issues of insufficient accuracy and DoT faced by LLMs-based agents in complex scientific reasoning tasks. As illustrated in Figure~\ref{figure1}, The RR-MP framework employs a multi-path reasoning mechanism, analogous to human reasoning—complex reasoning tasks typically require multiple reasoning paths to arrive at correct answers~\cite{stanovich2000advancing}. In each path, to enhance reasoning accuracy, we optimize each pathway through iterative reflection, thereby preventing the occurrence of DoT during the iterative process. The framework integrates reactive agents and reflection agents working collaboratively; the reflection agents stimulate reactive agents to perform self-correction. The dual-system model, comprising reactive agents and reflection agents, is reminiscent of the two systems in human cognition—System 1 (fast and intuitive) and System 2 (slow and deliberative)—thereby effectively enhancing decision-making performance~\cite{kahneman2011thinking, vygotsky1978mind, christakopoulou2024agents}. We validated our approach in zero-shot and few-shot scenarios across three complex scientific reasoning tasks—moral scenarios, college physics, and college mathematics. The results indicate that self-correction through external stimulation and optimization of reasoning paths achieves higher accuracy. Notably, our method significantly outperforms strong baseline methods in zero-shot settings~\cite{brown2020language, wei2022chain, ma2023let, wang2022self,madaan2024self}. 

In summary, our contributions are summarized as follows:
\begin{itemize}
    \item We propose the Reactive and Reflection agents with Multi-Path Reasoning (RR-MP) framework as an effective post-hoc error-correction approach, aimed at significantly enhancing agents’ reasoning capabilities in complex scientific tasks.

    \item We propose a multi-path reasoning mechanism that enables multiple reactive agents to generate parallel reasoning paths, thereby improving accuracy and robustness in complex scientific reasoning.

    \item We conduct a comparative analysis of the performance of reactive and reflection agents under various prompt types, and further investigate how different communication modes (e.g., collaboration and debate) influence scientific reasoning outcomes.
\end{itemize}

\begin{figure*}[t]
\centering
\includegraphics[width=1\textwidth]{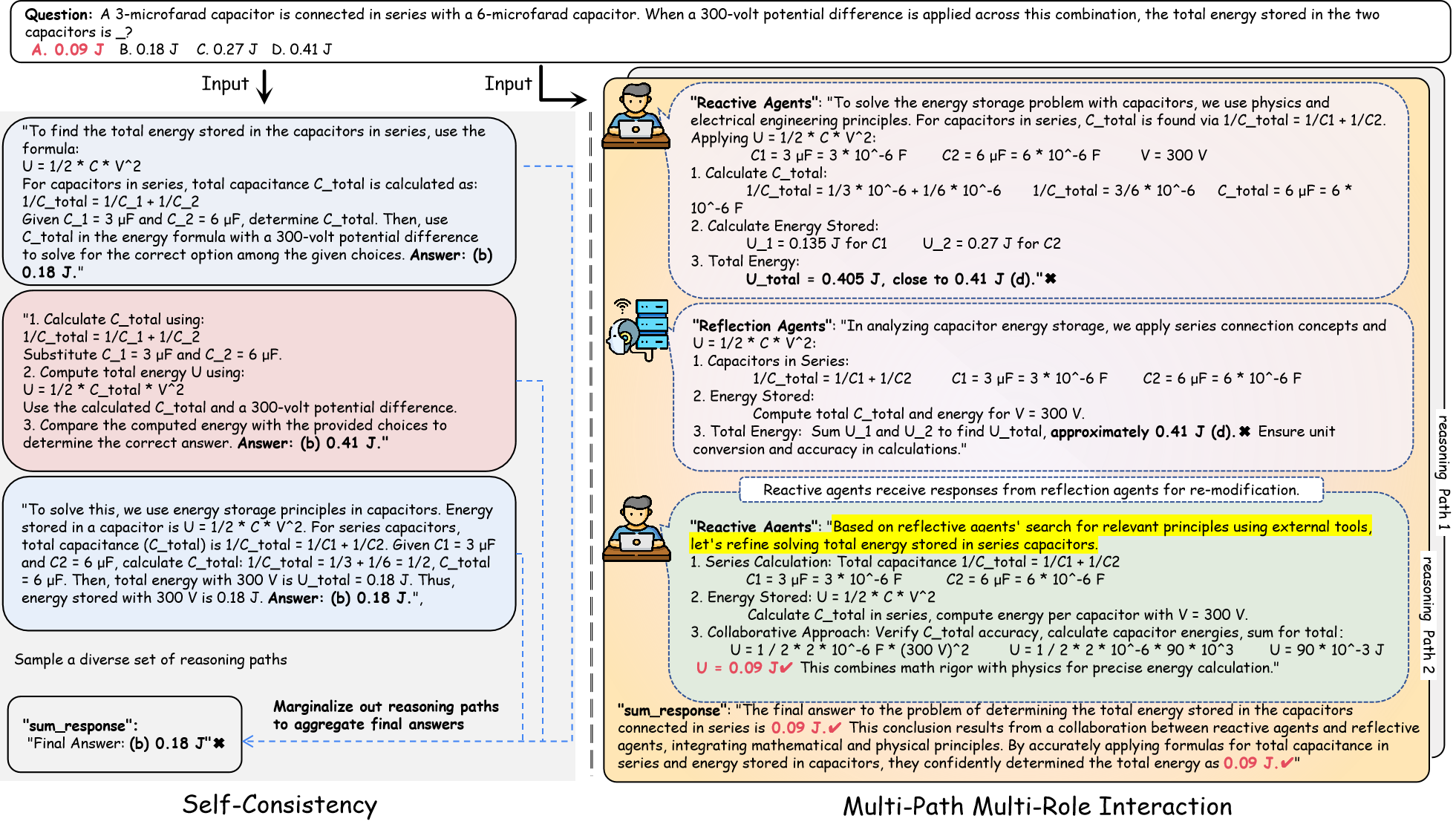} 
\caption{\fontfamily{ptm}\selectfont A comparison of the self-consistency method and our approach. Using the College Physics datasets as an example, our multi-path, multi-role interactive framework effectively mitigates errors caused by the majority of incorrect judgments in majority voting and leverages accurate reflection stimulated by the reactive agents' input on the reflection agents' reasoning. Even if the first path yields an incorrect result, the final answer is achieved through reflection analysis of the second path. Refer to B.1 in the appendix for details.}
\label{figure2} 
\end{figure*}

\section{Related Work}

\textbf{Self-Correction of a Single agent.} 

Current LLMs still exhibit limitations in scientific reasoning, with accuracy often compromised due to hallucinations. Developing a simple yet effective approach to enhance the self-correction capabilities of intelligent agents remains a critical challenge~\cite{saunders2022self, chen2023iterative, feng2024large, wang2023aligning, yax2024studying}. \citet{wei2022chain} proposed the chain-of-thought method, which improves the model's complex reasoning ability through intermediate reasoning steps. Additionally, researchers have suggested decomposing complex problems into simpler subproblems to enable LLMs to plan in a manner similar to the human brain~\cite{hao2023reasoning, zhou2022least, khot2020text}. These works lay the foundation for subsequent self-correction mechanisms. \citet{wang2022self} introduced self-consistency decoding, which addresses repetition and local optimum issues in chain-of-thought prompts, reducing randomness during the generation process. \citet{madaan2024self} proposed the SELF-REFINE method, where the agent first generates an output based on a given input and passes it back to the same model for feedback. The feedback is then returned to the model for optimization. This process iterates until a stopping condition is met. However, a single agent often lacks sufficient decision-making and planning abilities when dealing with complex tasks~\cite{ma2023let, sel2024skin}. One aspect of our work is to optimize iterative output and feedback among multiple agents, effectively avoiding the DoT that occurs during self-reflection in a single agent.

\textbf{Collaborative Error Correction in Multi-agent Systems.} The outputs of multi-agent systems can effectively correct errors, thereby enhancing the efficiency and accuracy of solving complex problems~\cite{guo2024large,zhu2024distilling}. A multi-agent system consists of multiple autonomous agents that interact with each other. By leveraging a shared environment or tasks, it facilitates the distributed resolution of decision-making problems. This collaborative approach can significantly improve the efficiency and accuracy of multi-agent systems in solving complex problems~\cite{rasal2024llm, zhu-etal-2023-solving, pan2024human, he2023solving}. For example, agents proficient in physical models can perform physical logical reasoning more effectively but are prone to calculation errors when dealing with formulas. In contrast, agents skilled in mathematical computations can reflect and correct the calculation structure of physical model agents, thereby solving complex university-level physics problems~\cite{chen2024comm}. Additionally, critical interactions among agents are another effective pathway to enhance the ability of multi-agent systems to solve complex problems. Related studies have shown that utilizing multiple agents for critical debates can enhance problem-solving capabilities and mitigate the DoT through debate~\cite{du2023improving, liang2023encouraging}. Multi-agent systems based on LLMs have already demonstrated encouraging collective intelligence. However, current multi-agent systems still face limitations in demand responsiveness, as tasks are often handled by fixed agents, and feedback mechanisms for intermediate tasks remain insufficient. These shortcomings restrict the adaptability and decision-making efficiency of multi-agent systems in complex scenarios. 

Our research is closely related to the field of multi-agent systems, with a focus on exploring the effectiveness of the RR-MP framework. We guide LLMls to generate diverse reasoning paths, simulating the human experience of observing the world from different reasoning perspectives to derive accurate answers~\cite{kahneman2011thinking}. This enables multiple agents to dynamically collaborate and achieve diversified demand responsiveness, thereby improving system performance. To address the issue of insufficient feedback mechanisms, we design an interaction framework between reactive and reflection agents, enhancing the timeliness and effectiveness of reasoning feedback through collaborative correction and information sharing. This method leverages the collaborative capabilities of agents to achieve efficient self-correction and optimization.

\section{Methods}

We introduces our proposed RR-MP framework, which is divided into two parts. Section~3.1, \textit{Multi-Path Reasoning for Enhanced Cognitive Flexibility}, demonstrates the effectiveness of multi-path reasoning through theoretical analysis. Section~3.2, \textit{Multi-agent Interactions for Collaborative Cognitive Task Solving}, describes the communication mechanisms between reactive and reflection agents and provides a detailed analysis in the experimental section.

\subsection{Multi-Path Reasoning for Enhanced Cognitive Flexibility}

We adopt a multi-path reasoning approach to emulate the collaborative behavior of human teams. Specifically, when different members of the team produce consistent answers, it increases our confidence in the correctness of the solution. Unlike self-consistent methods that rely on aggregating multiple reasoning paths to achieve consensus~\cite{wang2022self}, our approach not only integrates decision outcomes from multiple paths but also conducts in-depth analyses to derive the final decision. This enables the timely and effective evaluation and correction of the reasoning process, even when most initial paths are incorrect, potentially allowing the corrected answer to be output as the final result, as illustrated in Figure~\ref{figure2}.

The core of our RR-MP framework is to achieve optimal solutions through diverse reasoning pathways. We assigned specific roles to the agents in each reasoning path to encourage collaboration among agents with diverse roles to solve the target task. This approach represents a simple yet effective prompting technique~\cite{wang2023rolellm}, and our design principles follow those of~\citet{chen2024comm}, with detailed implementation provided in Appendix B. By leveraging multiple diverse and reasonable reasoning paths generated by different roles, we ultimately achieved the optimal solution. To validate this, we conducted a theoretical proof. Following \citet{sel2024skin}, we view the reasoning process in complex problem-solving as an implicit optimization  (a.k.a. mesa-optimization~\cite{hubinger2019risks}) of the overall welfare function contributed by multi-path reasoning roles. We now perform a theoretical analysis of multi-path reasoning, assuming we have a problem datasets $Q$, an action space $A$, and a prompt system $p$. For a single query $q \in Q$, there is a specific action decision $a \in A$ that yields the optimal $F^S(q)$. We can consider the decision process for $F^S(q)$ as an implicit optimization process, where the function $F^S(q)$ represents the decision function $F$ of the decision-maker $S$, who is responsible for making the final decision. We formalize this process as:

\begin{equation}
F^S(q) = \arg\max_{a \in A} \prod_{i=1}^n 
\mathbb{E}_{x \sim h^{\mathrm{m}_i}(q,F^{\mathrm{p}_i}(a))}
h_u^{\mathrm{p}_i}(x)
\end{equation}

where $h^m : Q \times A \to \mathcal{P}(\mathcal{X})$ serves as a logic generator within a multi-agent interaction in a multi-path Framework, inferring the logic of possible decision processes based on a given query $q \in Q$ and the prompt from the prompting system $p$. $\mathcal{P}(\mathcal{X})$ is the set of all probability distributions over the decision space $\mathcal{X}$. The term $\arg\max_{a \in A}$ represents maximizing the expected value of all paths under a specific action $a$. The symbol $\prod$ indicates the product over all possible paths, denoted by $\Pi^n$, where $i$ represents the $i$-th path in the set and $n$ is the total number of paths. The expectation operator $\mathbb{E}$ represents the expected value of the random variable $x$, and $x$ is the random variable representing outcomes generated by different reasoning logics. The notation $\sim$ signifies that the distribution of $x$ follows a probability distribution. The method $h^{\mathrm{m}}(q, F^{\mathrm{p_i}}(a))$ generates the random variable $x$ for the question $q$ using the decision $F^{\mathrm{p}}(a)$.  The term $F^{\mathrm{p_i}}(a)$ denotes the optimal decision for the question $q$ along the $i$-th path. The utility function $h_u^{\mathrm{p}}(x)$ represents the utility of the outcome $x$ along this path. Overall, $h$ represents the utility function, reflecting the effectiveness of the method, which is manifested in the correctness of the final answer. The symbol $\mathrm{p}$ denotes the specific path, and $u$ represents the overall utility or effectiveness value of the method.\\

We assume the utility function \( h_{u}^p(x) \) is consistent. Let \( X_1^{q,a}, \ldots, X_n^{q,a} \) be i.i.d. samples from the distribution \( h_{s}^p(q,a) \). The true utility \( G_p(x) \) we want to optimize through the prompt system \( p \) is consistent, i.e., \(\mathbb{E} \left[ G_p(x) \right] = \mathbb{E} \left[ \prod_{i=1}^n h_{u}^p(x_i) \right]\). Define the total variation distance between two distributions as \( D_{TV}(Z_1 \| Z_2) = \sup_{A \subseteq Z} |Z_1(A) - Z_2(A)| \).  We obtain the following inequality:
\begin{equation}
\begin{aligned}
P \Bigg( & \Bigg| \mathbb{E}_{x \sim h^{m_i}(q, F^{p_i}(q))} \prod_{i=1}^n h^{p_i}_u(x)\\
&- \mathbb{E} \left[ \frac{1}{n} \sum_{i=1}^n \prod_{j=1}^n h^{p_i}_u(X_i^{q,a}) \right] \Bigg| \geq t \Bigg) \leq \frac{\sigma^2}{n t^2}
\end{aligned}
\end{equation}
where, for any query \( q \in Q \), any decision \( a \in A \), and error bound \( t \in \mathbb{R}^+ \), can be defined as:
\begin{equation}
t = \left\| G \right\|_{\infty} D_{TV} \left[ X^{q,a} \big\| h^{\text{m}_{i}}(q,a) \right] + \epsilon
\end{equation}
where \( \|G\|_{\infty} \) provides the maximum oscillation range of \( G \) under all inputs. \( D_{TV} \left[ X^{q,a} \big\| h^{\text{m}_{i}}(q,a) \right] \) gives the maximum discrepancy between the empirical distribution of the samples and the theoretical distribution. \(\epsilon\) is a small adjustment parameter used to add extra tolerance for error.

Furthermore, we have the equation:
\begin{equation}
\begin{aligned}
P \Bigg( & \Bigg| \mathbb{E}_{x \sim h^{\mathrm{m}_i}(q, F^{\mathrm{p}_i}(q))} G_P(x) 
- \mathbb{E} \left[ \frac{1}{n} \sum_{i=1}^n \prod_{j=1}^n h^{\mathrm{p}_i}_u(x_i^{q,a}) \right] \Bigg| \\
& \geq \left\| G \right\|_{\infty} D_{TV} \left[ X^{q,a} \big\| h^{\text{m}_{i}}(q,a) \right] + \epsilon \Bigg) \leq \frac{\sigma^2}{n t^2}
\end{aligned}
\end{equation}

Based on Chebyshev's inequality, as \( n \) increases, the probability that the deviation exceeds a fixed value \( t \) decreases, which means the probability of an error occurring decreases. The formula is as follows:
\begin{equation}
\begin{aligned}
\Bigg| \mathbb{E}_{x \sim h^{m_i}(q, F^{p_i}(q))} G_P(x)
& - \mathbb{E} \left[ \frac{1}{n} \sum_{i=1}^n \prod_{i=1}^n h^{p_i}_u(x_i^{q,a}) \right] \Bigg| \\
& \rightarrow 0 \quad \text{as} \quad n \rightarrow \infty
\end{aligned}
\end{equation}

Therefore, we conclude that under the given assumptions, the optimization result of the formula \( G_p(q) \) is proven through the aforementioned inequality. This demonstrates that by combining the expected utilities of different agents' paths and methods, we can identify the optimal decision that maximizes the utility function. Furthermore, we theoretically prove that as the number of agents increases, the generated multi-path reasoning significantly enhances decision quality. This conclusion is consistent with the experimental results of \citet{wang2022self}.

\begin{figure*}[t]
\centering
\includegraphics[width=\textwidth]{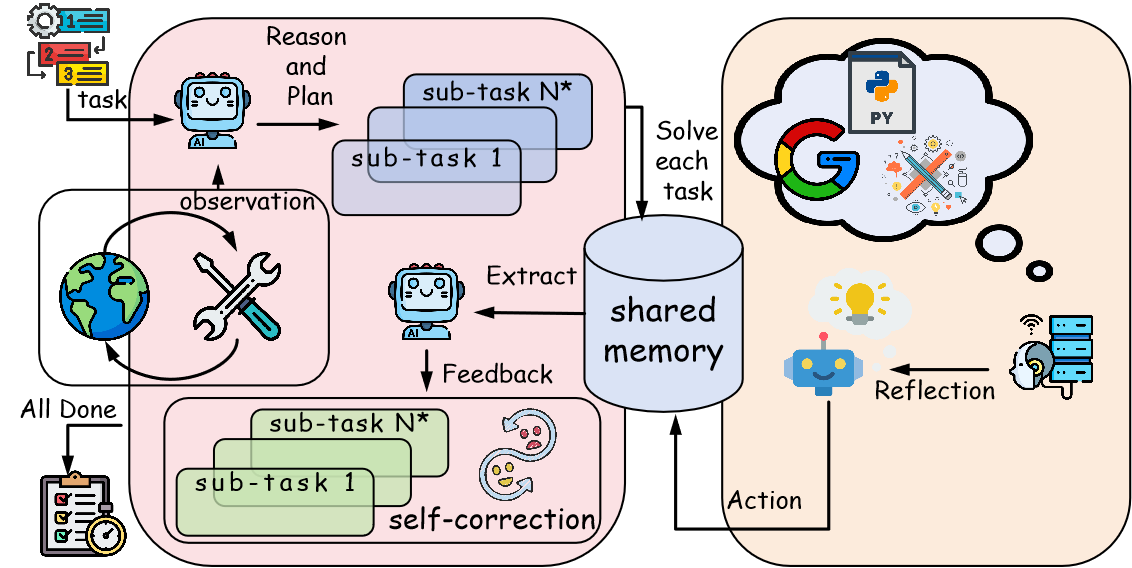} 
\caption{The reasoning process of the reactive agent and reflection agent. The reactive agent receives information from the external environment, decomposes it into sub-tasks, and stores them in the database. The reflection agent performs each sub-task through a process of supplementation or critique and returns the results to the reactive agent. Based on the feedback, the reactive agent refines its reasoning and completes the scientific reasoning process.}
\label{figure3}
\end{figure*}

\subsection{Multi-agent Interactions for Collaborative Cognitive Task Solving}

In this chapter, we introduce the interaction process between reactive agents and reflection agents within a specific path of a multi-path reasoning framework. As illustrated in Figure~\ref{figure3}, the primary interaction between the reactive agent and the reflection agent occurs through the shared memory module Shared memory (retrieved and stored in list format). The preliminary responses generated by the reactive agent are stored in shared memory , from which the reflection agent retrieves these responses for further analysis and processing. Before the final answer is obtained, the reactive agent awaits the completion of the reflection agent's analysis until the final answer is generated. The following sections provide a detailed description of this process.

The reactive agent maintains a partially observable understanding of the environment. Upon receiving a question datasets $Q$, it generates a specific action decision $a'$. Through its actions, the reactive agent produces a preliminary answer $s'$ to address the problem, which is stored in memory as a dictionary entry, awaiting extraction by the reflection agent. Once the reflection agent retrieves the preliminary answer $s'$ from the reactive agent, it undergoes multi-step reasoning and, with the assistance of relevant tools and external knowledge, formulates the reasoning strategy $\pi$.

The language-based agent selects and extends the initial action $a'$ based on the strategy $\pi$ implemented by a LLM with parameters $\Theta$, adhering to a set of instructions $p$ provided via prompts. The reflection agent's inputs include the instructions $p$, the preliminary response $s'$, and the original question $Q$. We formalize this process as follows: during the update phase, the language-based agent selects an action $a \in A$ based on the strategy $\pi$ implemented by the LLM with parameters $\Theta$:

\begin{equation}
    a\sim\pi(a^{\prime}|p,s^{\prime};\Theta)
\end{equation}

Consequently, after the interaction between the reactive agent and the reflection agent, the original action decision $a'$ is expanded to action $a$, a process referred to as an ``augmented action''. By partially observing the task information $b'$, and utilizing the LLM with parameters $\Theta$ to invoke tools or obtain information from external knowledge bases, and under the constraints of instructions $p$ to formulate the final strategy $\pi$, the newly augmented action $a$ is executed. This process effectively enhances decision-making performance.

\section{Experiments}
\subsection{datasets and Baseline Methods}

We select three datasets—College Physics, College Mathematics, and Moral Scenarios—from the Massive Multitask Language Understanding (MMLU) benchmark~\cite{hendrycks2020measuring} to evaluate the performance of large language models in scientific reasoning tasks. The College Physics datasets evaluates mastery of domain-specific physical knowledge, the College Mathematics datasets focuses on logical reasoning and complex computation, and the Moral Scenarios datasets examines ethical decision-making and abstract reasoning. Together, these datasets capture the core requirements of scientific reasoning tasks and present significant challenges to large language models~\cite{ma2023let,chen2024comm}. With its broad coverage of key areas in scientific reasoning, the MMLU benchmark serves as a powerful tool for identifying model blind spots in domain knowledge, causal reasoning, and value-based judgment, offering a comprehensive evaluation of reasoning capabilities.

To evaluate the effectiveness of the \textit{Reactive and Reflection agents with Multi-Path Reasoning} method in scientific reasoning tasks, we compared five baseline methods in zero-shot and few-shot settings. Each method represents a different paradigm of reasoning and decision-making for agents, as detailed below:
\begin{enumerate}
    \item \textbf{Standard}~\cite{brown2020language}: This method simulates the traditional approach where the agent directly generates an output from the input without engaging in any reasoning or self-reflection. It is suitable for tasks that prioritize efficiency.
    \item \textbf{Chain-of-Thought (CoT)}~\cite{wei2022chain}: In this method, the agent performs step-by-step reasoning before making a decision and provides a detailed explanation of its reasoning process. This approach is particularly effective for complex decision-making tasks and mimics the human process of breaking down problems into sequential steps.
    \item \textbf{Thought Experiment (Thought)}~\cite{ma2023let}: This method involves counterfactual reasoning, where the agent considers various (often hypothetical) scenarios and carefully analyzes the potential outcomes of these imagined situations, supporting more comprehensive decision-making.
    \item \textbf{Self-Consistency}~\cite{wang2022self}: Instead of relying on a single greedy reasoning path, this method samples multiple reasoning paths. The final answer is determined by marginalizing over the sampled reasoning paths to select the most consistent solution.
    \item \textbf{Self-Refine}~\cite{madaan2024self}: This method is based on large language models (LLMs) and focuses on iterative self-improvement. The agent generates an initial output and then provides feedback on its own output, iteratively refining it to produce a more accurate result.
\end{enumerate}

\subsection{Settings}

Due to resource limitations, we selected "gpt-3.5-turbo-0613" as the backbone model for all experiments. In our RR-MP framework, we designed an interaction framework between the reactive agent and the reflection agent. The reactive agent receives inputs from datasets, including College Physics, College Mathematics, and Moral Scenarios, makes initial decisions, and passes them to the reflection agent. The reflection agent further refines and optimizes these initial decisions through collaboration and debate, ensuring their accuracy and rationale. The two agents act as the "initial decision-maker" and the "decision optimizer," respectively, working together to complete tasks.

We tested the system in zero-shot and few-shot settings. In the zero-shot setting, the model relies entirely on its reasoning ability to make decisions without any prior examples. In the few-shot setting, five learning examples were provided for each agent to help them better understand the task context and decision-making logic. To further enrich the reasoning paths, we adopted a role-playing approach. For example, in the College Physics experiments, roles such as physicists and mathematicians were defined (based on simple prompt engineering). These roles, following design principles~\cite{chen2024comm}, explore diverse reasoning paths~\cite{wang2023rolellm}, with each role assuming specific tasks during the reasoning process and contributing to decision-making. The details of role definitions, task assignments, and the implementation of prompt engineering are thoroughly described in Appendix B.

\begin{table*}[t]
  \begin{center}
    \resizebox{\textwidth}{!}{ 
    \begin{tabular}{@{}lccccccc@{}}
      \toprule
      \multirow{2}{*}{Method} & \multicolumn{3}{c}{Zero-shot} & \multicolumn{3}{c}{Few-shot} & \multirow{2}{*}{Average} \\
      \cmidrule(lr){2-4} \cmidrule(lr){5-7}
      & Moral Scenarios & College Physics & College Math & Moral Scenarios & College Physics & College Math \\ 
      \midrule
      Standard~\cite{brown2020language} & 37.65 & 40.19 & 40 & 46.25 & 46.09 & 41 & 41.86 \\
      CoT~\cite{wei2022chain} & 48.49 & 57.84 & 39 & 52.29 & 63.72 & 38 & 48.22 \\
      Thought~\cite{ma2023let} & 41.45 & - & - & 49.5 & - & - & 45.48 \\
      Self-Consistency~\cite{wang2022self} & \underline{63.24} & 65.68 & 53 & \textbf{68.49} & 62.75 & 53 & 61.03 \\
      Self-Refine~\cite{madaan2024self} & 59.66 & 61.76 & 50 & 67.01 & 66.67 & 45 & 58.35 \\
      \midrule
      Same-Domain Collaboration & \textbf{70.39} & 85.29 & \underline{71} & 63.91 & 86.27 & \textbf{75} & \underline{75.15} \\
      Same-Domain Debate & 48.71 & \underline{87.25} & 70 & 62.12 & \underline{87.25} & \textbf{74} & 71.55 \\
      \cmidrule(lr){2-7}
      Different-Domain Collaboration & 60.78 & \textbf{89.21} & \textbf{74} & \underline{65.47} & \textbf{91.18} & \textbf{75} & \textbf{75.94} \\
      Different-Domain Debate & 59.77 & 85.29 & 74 & 56.76 & 86.27 & 70 & 72.02 \\
      \bottomrule
    \end{tabular}
    }
    \caption{Main Results (Accuracy, \%). ``Same-Domain Collaboration" indicates that the reactive agent and reflection agent collaborate within the same domain to perform scientific reasoning, while ``Different-Domain Debate" means they engage in debate across different domains. In the averages column, bold denotes the best result, and underline denotes the second-best result.}
    \label{tab1}
  \end{center}
\end{table*}

\subsection{Main Results}

In the proposed RR-MP framework, we designed four interaction paradigms to investigate the interplay between reactive agents and reflection agents: the first is collaborative interaction between the reactive agent and reflection agent in a same-domain context; the second is debate interaction in a same-domain context; the third is collaborative interaction between the two agents in a different-domain context; and the fourth is debate interaction in a different-domain context. The comparison results with baseline methods are shown in Table~\ref{tab1}, demonstrating that our approach achieves significant performance improvements in few-shot scenarios.

From the results in Table~\ref{tab1}, it can be observed that the RR-MP framework exhibits significant performance improvements under the human-machine collaboration paradigm across complex datasets in both zero-shot and few-shot scenarios, including College Physics, College Mathematics, and Moral Scenarios. Notably, the collaboration between reactive agents and reflection agents from different domains (Different-Domain Collaboration) achieves the best performance in the majority of tasks, with an average accuracy of 75.94\%, outperforming other baseline methods.

Furthermore, Table~\ref{tab1} reveals additional insights. For instance, collaboration modes generally outperform debate modes regardless of whether the agents are within the same domain or across different domains. This trend is consistently observed across multiple tasks in both zero-shot and few-shot settings. The collaboration mode aims to solve problems or reach consensus, enabling the integration of diverse perspectives, fostering a more comprehensive understanding, identifying blind spots, and preventing cognitive rigidity caused by debates. The study also finds that using reactive agents and reflection agents of the same type may lead to decreased performance when performing tasks. This is because when multiple agents of the same role collaborate, their thinking patterns and methodologies tend to converge, reducing diversity and innovation, thereby limiting performance on complex tasks.

In summary, the RR-MP framework significantly enhances the performance of complex reasoning tasks by designing flexible collaboration and debate modes and leveraging the diverse roles of reactive agents and reflection agents. The collaboration mode performs better in most scenarios, especially when integrating knowledge from different domains. Additionally, collaboration within the same domain can effectively facilitate task completion in specific tasks. These results validate the importance of multi-agent interaction design and provide strong support for the optimization of future multi-domain collaboration systems.

\begin{figure*}[t]
\centering
\includegraphics[width=\textwidth]{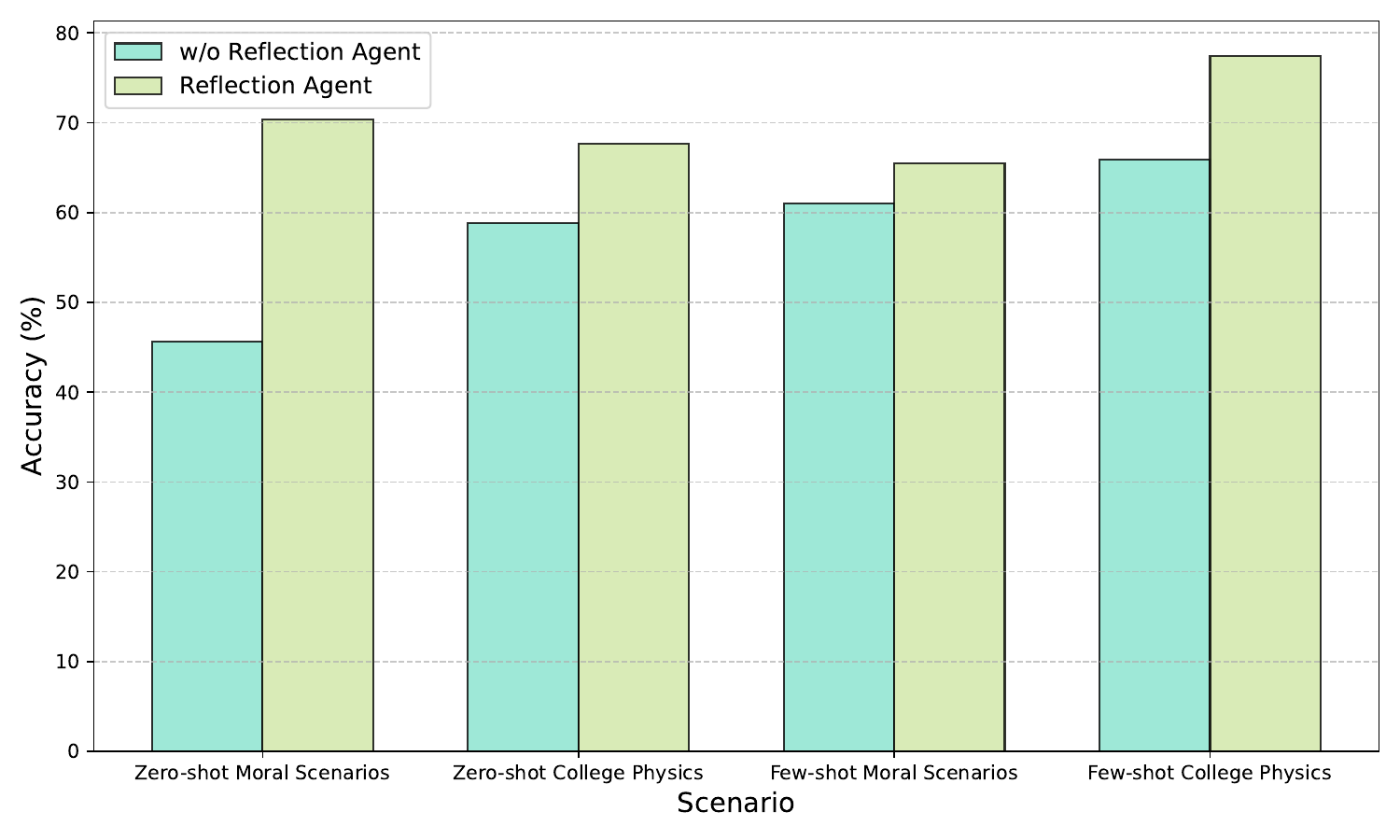} 
\caption{Accuracy (\%) with and without stimulation roles.}
\label{figure4}
\end{figure*}

\section{Ablation Study}

\begin{table}[t]
\centering
\resizebox{\columnwidth}{!}{%
\begin{tabular}{@{}lcccccc@{}}
\toprule
\textbf{Mode}                & \multicolumn{2}{c}{\textbf{College Physics (0-shot)}} & \multicolumn{2}{c}{\textbf{College Math (0-shot)}} & \multicolumn{2}{c}{\textbf{College Physics (few-shot)}} \\ 
\cmidrule(lr){2-3} \cmidrule(lr){4-5} \cmidrule(lr){6-7}
                             & \textbf{Single} & \textbf{Multiple} & \textbf{Single} & \textbf{Multiple} & \textbf{Single} & \textbf{Multiple} \\ 
\midrule
Same-Domain Collaboration      & 78.43           & \textbf{85.29}    & 69              & \textbf{71}       & 79.41           & \textbf{86.27}    \\ 
Same-Domain Debate             & 86.27           & \textbf{87.25}    & 67              & \textbf{70}       & 89.11           & 87.25             \\ 
Different-Domain Collaboration & 85.29           & \textbf{89.21}    & 71              & \textbf{74}       & 85.29           & \textbf{91.18}    \\ 
Different-Domain Debate        & 83.30           & \textbf{85.29}    & 70              & \textbf{74}       & 84.31           & \textbf{86.27}             \\ 
\bottomrule
\end{tabular}%
}
\caption{Performance comparison between single and multiple instances across different collaboration and debate modes.}
\label{tab2}
\end{table}

\begin{table*}[t]
  \begin{center}
    \fontsize{10}{12}\selectfont 
    \setlength{\tabcolsep}{1mm} 
    \begin{tabular}{lccc} 
      \toprule
      Method & Zero-shot Accuracy (\%) & Few-shot Accuracy (\%) & Average Accuracy (\%)\\
      \midrule
      Linear      & 59.00 & 53.90 & 56.45 \\
      Hierarchical & 63.72 & 57.80 & 60.76 \\
      Network  & 50.98 & 58.80 & 54.89 \\
      Ours         & \textbf{89.21} & \textbf{91.18} & \textbf{90.20} \\
      \bottomrule
    \end{tabular}
    \caption{Comparison of accuracy across three agents interaction methods  and our proposed RR-MP framework Results are evaluated on zero-shot, few-shot, and average accuracy (\%).}
    \label{tab3}
  \end{center}
\end{table*}

\subsection{Is It Necessary for reflection to Exist?}

In our proposed method, the reflection agent serves as a core component of the RR-MP Framework. We posit that the reflection agent plays a crucial role in exploring reasoning pathways during the reflection phase, particularly when the reactive agent exhibits hallucinations or overconfidence in its reasoning. In such scenarios, the reflection agent facilitates further cognitive optimization, analogous to how humans rely on external stimuli to refine their thought processes after encountering overconfident errors. To validate this hypothesis, we designed comparative experiments to assess the difference in reasoning performance between models with and without the reflection agent. Under both zero-shot and few-shot prompting settings, we conducted reasoning tasks on the Moral Scenarios and College Physics datasets, respectively. 

The experimental results are presented in Figure~\ref{figure4}. Specifically, under the zero-shot prompting setting, the reasoning accuracy on the Moral Scenarios and College Physics datasets improved by 24.81\% and 8.78\%, respectively. Under the few-shot prompting setting, the accuracy increased by 4.44\% and 11.55\%, respectively. These results indicate that incorporating the reflection agent significantly enhances the model's ability to handle complex reasoning tasks within these datasets. By introducing external stimulation to optimize reasoning pathways, the reflection agent can correct and augment cognitive processes, thereby ultimately achieving superior decision-making performance.

\begin{figure}[t]
    \centering
    \includegraphics[width=0.7\columnwidth]{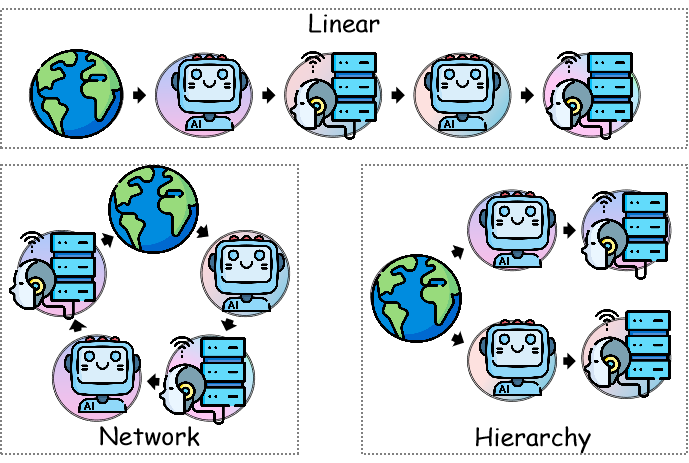}
    \caption{\fontfamily{ptm}\selectfont Three typical interaction paradigms in human-agent collaboration frameworks.}
    \label{figure5}
\end{figure}

\subsection{Are Multiple Instances Necessary?}

In our proposed method, the reactive agent and reflection agent are both based on the same type of large language models (LLMs), such as ChatGPT-3.5, but operate as independent dialogue instances without interference. This design ensures that each agent can perform its designated tasks independently, avoiding reasoning biases caused by shared context or cross-agent interference. To investigate whether it is possible to achieve multi-path reasoning by dynamically switching agents within a single-instance LLM through prompt engineering, we devised an alternative approach. This method simulates different agents within the same dialogue instance using prompt engineering and conducts four types of interaction experiments on the College Physics and Moral Scenarios datasets, as shown in Table~\ref{tab2}.

The experimental results show that, regardless of zero-shot or few-shot prompting settings, the reasoning performance of single-instance dialogues decreases. This decline can be attributed to context conflicts or inconsistencies arising from frequent switching between agent modes, which negatively impact prediction accuracy. In contrast, multi-instance dialogues maintain consistency and independence among agents, significantly enhancing collaboration and improving reasoning performance. Moreover, single-instance dialogues are more costly, as they require frequent input of role-specific information, whereas multi-instance setups only require a single input for each agent.Our findings align with the studies of~\citet{xu2023expertprompting, chen2024comm}, which emphasize the importance of clear definitions and independent task boundaries for each agent. Well-defined agent roles not only help maintain the self-consistency of LLMs but also effectively prevent cognitive confusion, thereby improving response quality and the reasoning ability to address complex scientific problems.

\subsection{Exploring the Impact of Interaction Methods on agents.}

In our study, we introduced three typical topological structures to explore interaction strategies in multi-agent systems: Linear Interaction, Network Interaction, and Hierarchical Interaction, as shown in Figure~\ref{figure5}. These interaction methods are inspired by common patterns of human team collaboration. Linear Interaction is a sequential approach where agents process and transfer tasks along a fixed linear path, resembling workflows in assembly lines or hierarchical organizations. Network Interaction allows agents to establish arbitrary dependencies within a networked structure, reflecting the flexibility and dynamic adjustments often observed in team-based collaboration. Hierarchical Interaction adopts a layered structure where agents work in parallel across different branches, similar to team collaboration based on roles or functional hierarchies.

We conducted tests on the College Physics datasets. Experimental results (Table~\ref{tab3}) demonstrate that, while Hierarchical Interaction exhibits relatively well performance, our proposed RR-MP framework achieves significantly better results due to its reflection capability. reflection enables agents to dynamically adjust reasoning paths during interactions, effectively enhancing their ability to self-correct and optimize when addressing complex scientific problems. By combining reflection with multi-path reasoning, our method exhibits superior flexibility and efficiency across all scenarios, further validating the importance of reflection and dynamic interactions in the design of multi-agent systems.

\section{Conclusion}

In this work, we propose a framework named Reactive and Reflection agents with Multi-Path Reasoning. This framework aims to address the issues of decreased accuracy and Degeneration-of-Thought in multi-agent systems during complex scientific reasoning, which are caused by fixed single responses and insufficient execution of intermediate feedback. By doing so, it enhances the reasoning capabilities of LLMs in solving complex scientific problems. Our approach consists of two core components: first, the diversity of multi-path reasoning methods significantly improves the accuracy of LLMs; second, the interaction between multiple agents effectively mitigates hallucinations and Degeneration-of-Thought issues. We have demonstrated the effectiveness of the framework through both theoretical analysis and experimental validation.

Although the proposed framework serves as an effective post-hoc error correction method, significantly improving the decision-making capabilities of agents in complex tasks, it still has certain limitations. Specifically, the framework requires task-specific design of roles and reasoning examples, which is a common challenge in the field of prompt engineering~\cite{wei2022chain,ma2023let,liang2023encouraging,madaan2024self}. Future work will focus on exploring how to implement automated prompt design within the framework to further enhance the method's generalizability and adaptability.

\bibliographystyle{elsarticle-num-names}
\bibliography{References}
\end{document}